\documentclass[conference]{IEEEtran}
\usepackage{cite}
\usepackage{amsmath,amssymb,amsfonts}
\usepackage{algorithmic}
\usepackage{graphicx}
\usepackage{textcomp}
\usepackage{xcolor}
\def\BibTeX{{\rm B\kern-.05em{\sc i\kern-.025em b}\kern-.08em
    T\kern-.1667em\lower.7ex\hbox{E}\kern-.125emX}}

\usepackage[numbers,sort&compress]{natbib}
\usepackage{subfigure}
\usepackage{wrapfig}
\usepackage{multirow}
\usepackage{booktabs}       
\usepackage{hyperref}

\usepackage[linesnumbered,ruled,vlined,noend]{algorithm2e}
\usepackage{amsmath}
\usepackage{amssymb}
\usepackage{amsthm}
\usepackage{bm}
\usepackage{bbm}
\usepackage{enumitem}
\usepackage{tikz}
\usetikzlibrary{backgrounds}
\usetikzlibrary{bayesnet}
\usetikzlibrary{arrows}

\newcommand{\A}{\bm{A}}

\newcommand{\ba}{\bm{a}}

\newcommand{\EX}{\mathbb{E}}

\newcommand{\testy}{y_0}
\newcommand{\mygraphwidth}{0.6\linewidth}
\newcommand{\myiconwidth}{0.35\linewidth}
\newcommand{\myiconsubheight}{0.5\linewidth}

\begin{document}

\title{Conditional Imitation Learning\\ for Multi-Agent Games}

\author{
\IEEEauthorblockN{Andy Shih}
\IEEEauthorblockA{\textit{Computer Science Department} \\
\textit{Stanford University}\\
andyshih@cs.stanford.edu}
\and
\IEEEauthorblockN{Stefano Ermon}
\IEEEauthorblockA{\textit{Computer Science Department} \\
\textit{Stanford University}\\
ermon@cs.stanford.edu}
\and
\IEEEauthorblockN{Dorsa Sadigh}
\IEEEauthorblockA{\textit{Computer Science Department} \\
\textit{Stanford University}\\
dorsa@cs.stanford.edu}
}

\maketitle

\begin{abstract}
While advances in multi-agent learning have enabled the training of increasingly complex agents, most existing techniques produce a final policy that is not designed to adapt to a new partner's strategy.
However, we would like our AI agents to adjust their strategy based on the strategies of those around them. 
In this work, we study the problem of conditional multi-agent imitation learning, where we have access to joint trajectory demonstrations at training time, and we must interact with and adapt to new partners at test time. This setting is challenging because we must infer a new partner's strategy and adapt our policy to that strategy, all without knowledge of the environment reward or dynamics.
We formalize this problem of conditional multi-agent imitation learning, and propose a novel approach to address the difficulties of scalability and data scarcity.
Our key insight is that variations across partners in multi-agent games are often highly structured, and can be represented via a low-rank subspace. 
Leveraging tools from tensor decomposition, our model learns a low-rank subspace over ego and partner agent strategies, then infers and adapts to a new partner strategy by interpolating in the subspace.
We experiments with a mix of collaborative tasks, including bandits, particle, and Hanabi environments. Additionally, we test our conditional policies against real human partners in a user study on the Overcooked game. Our model adapts better to new partners compared to baselines, and robustly handles diverse settings ranging from discrete/continuous actions and static/online evaluation with AI/human partners.
\end{abstract}


\section{Introduction}

Many important robotics applications naturally involve multiple agents, from assistive robotics to self-driving cars. New techniques in deep multi-agent reinforcement learning have led to breakthrough performance in many multi-agent tasks, such as Go~\cite{silver2016mastering}, Hanabi~\cite{Foerster2018BayesianAD}, and poker~\cite{brown2019superhuman}. Although these methods have shown impressive results, many of their formulations lack a key factor that is central to multi-agent interactions -- the ability to adapt quickly to another agent.

For example, in cooperative games such as Hanabi~\cite{Foerster2018BayesianAD,hu2019simplified,bard2020hanabi}, much of the focus has been on training a single set of partners to achieve a high score with each other. As a result, these methods produce agents that are skilled not at playing Hanabi in general, but at playing Hanabi with their training partners. Even in competitive or mixed settings, most current frameworks do not act with the
opponent in mind~\cite{brown2019superhuman, paquette2019diplomacy}. For example, state-of-the-art poker agents “play a fixed strategy that does not adapt to the observed tendencies of the opponents”~\cite{brown2019superhuman}. Recent works on the game of Diplomacy consider the ``exploitability’’ of their own agent (e.g. if adaptive opponents can take advantage of their agent), but do not adapt to or exploit their opponents’ behavioral patterns in return~\cite{gray2021humanlevel}.

\begin{figure*}[t]
    \begin{minipage}{0.65\linewidth}
        \centering
        \includegraphics[width=1.0\linewidth]{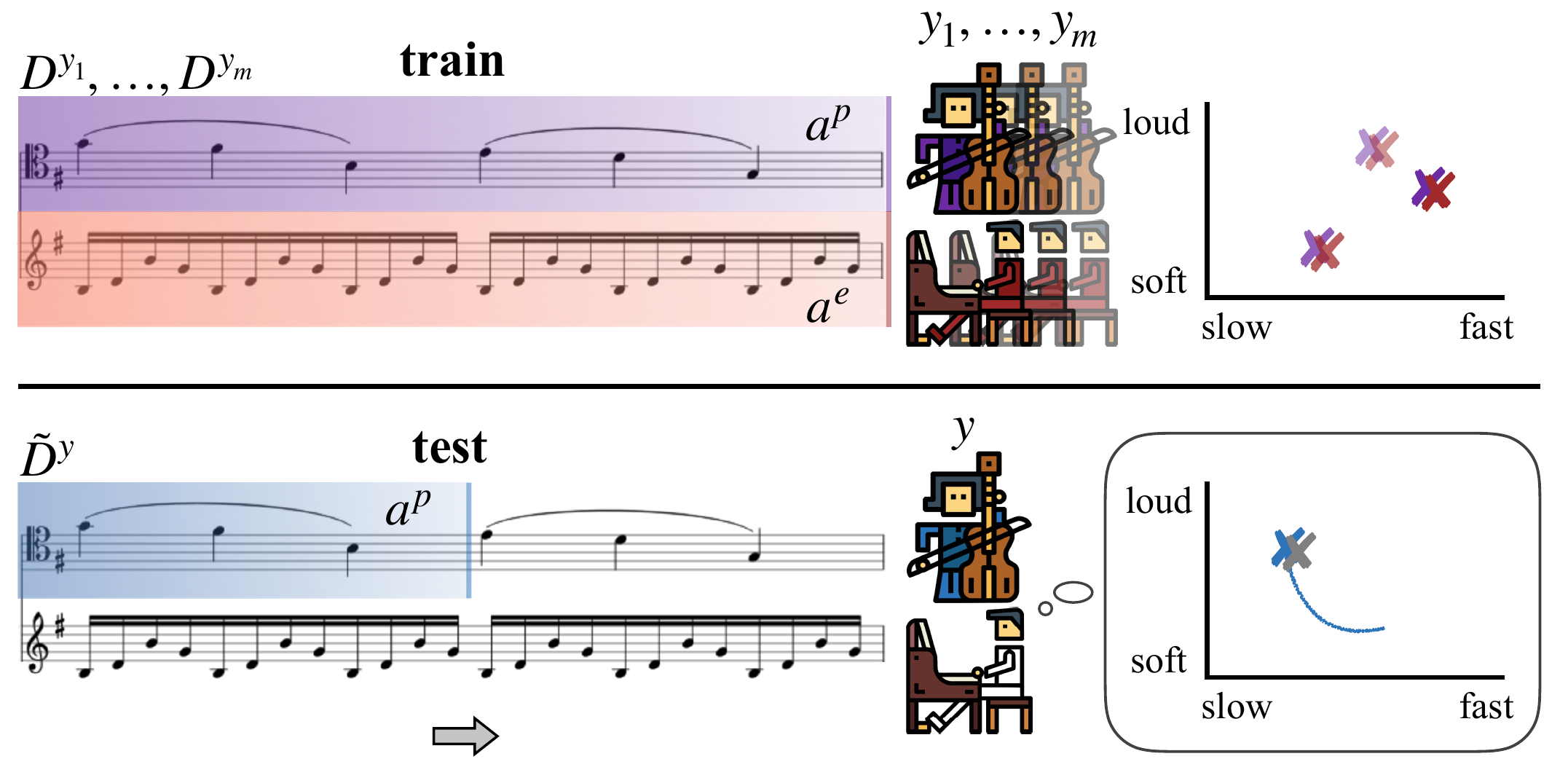}
    \end{minipage}
    \hfill
    \begin{minipage}{0.32\linewidth}
        \centering
        \caption{We consider the setting of adapting to a new partner in a multi-agent task. During training, the ego-agent (pianist in white) learns from demonstrations from pairs of cellists and pianists playing music in coordination. We denote each cellist/pianist pair with a different shade of purple/red, with the cellists representing the partners, and the pianists representing the experts. At test time, the ego-agent must coordinate with the cellist \(y\) in blue. To coordinate well, the pianist in white can first build a mental model from the training data of how to coordinate with partners of various strategy (loud/soft, fast/slow). Then, he can infer the strategy of the new partner in blue, and correctly accompany her playing style.}
        \label{fig:cover}
    \end{minipage}
\end{figure*}

Rather, success in a multi-agent task should be measured in terms of the ability to perform well at the task with a new partner. Indeed, a skillful musician can adjust to the playing style of new partners (Figure~\ref{fig:cover}), and a skillful poker player can exploit the bluffing patterns of new opponents. Similarly, we should train agents that are able to adapt their actions based on the tendencies of new partners.

In this work, we explore the paradigm of conditional multi-agent imitation learning (conditional MAIL). Our goal is to learn a policy that can adapt to new partners, by training only on a dataset of demonstrations without any other assumptions such as access to environment reward or dynamics. Concretely, we are provided with expert demonstrations on how to coordinate with various partners sampled from some fixed partner distribution. Then, given the actions of new partners, we would like to adapt our actions to best coordinate with them.

Conditional MAIL is a flexible framework, but there are two core challenges: 1) multi-agent evaluation and 2) data scarcity. 
\subsubsection{Evaluation}
Unlike the single-agent case where we can accurately estimate the performance of the learned policy, the learned policy performance for the multi-agent setting depends highly on the partner's strategy and the potential non-stationarity of that policy.

To look at trade-offs of evaluation cost and effectiveness, we experiment with three evaluation methods in our paper: \emph{offline}, \emph{static} and \emph{online evaluation}. \emph{Offline evaluation} simply measures the log-likelihood of a held out expert dataset as a proxy for reward, and can be evaluated without access to the environment even at test time. Another direct approach, which we refer to as \emph{static evaluation}, is to behavioral-clone (BC) a test dataset of trajectories, and evaluate our ego agent against these BC agents. The downside of static evaluation is that BC agents have rigid behavior and work poorly if out of distribution. This is exacerbated in multi-agent games, where other agents can quickly push the joint state out of distribution.

A more costly evaluation choice is \emph{online evaluation} by pairing our conditional policy against non-stationary partners, such as human partners. \emph{Online evaluation} is more meaningful but also more tedious. A highlight of our work is the ability of our policy to interact and adapt to human users in an online fashion, demonstrated on the game of Overcooked. Our policies finetune their behavior after each episode of interaction with the user.

\subsubsection{Data Scarcity}
Another challenge with conditional multi-agent imitation learning is data scarcity. To learn a conditional policy, we need training data consisting of a diverse set of coordination behavior, which is often hard to come by in practice. For example, the Overcooked dataset~\cite{Carroll2019OnTU} only has around \(15\) joint trajectories per game layout. A limited dataset, combined with the lack of access to environment dynamics for simulation, means that our algorithms must learn an adaptable latent space using only a handful of demonstrations.

To this end, our paper proposes a novel approach of synthesizing a low-rank space of policies via tensor decomposition. Learning a low-rank policy space enables us to scale to high dimensional environments, while avoiding the inherent difficulties of learning a non-linear latent space on limited data.
Inspired by works that model the value function of reinforcement learning tasks with low-rank decomposition~\cite{gorodetsky2018high,Yang2020Harnessing} and works that model human preferences in recommendation systems also with low-rank decomposition~\cite{schafer2007collaborative,he2017neural,yin2021ttrec}, we hypothesize that policies of either AI or human partners can admit a low-rank structure as well.
We use techniques from tensor decomposition, in particular Tensor Trains, to model this low-rank representation in a scalable way. This formulation then allows us to adapt quickly to a new partner by interpolating in the low-rank subspace. 

\subsection{Contributions}
The contributions of our paper are three-fold. First, our paper examines the various components (e.g. training objective, evaluation metric, user study integration) of the framework of conditional multi-agent imitation learning. 
Second, we propose a model  with a structured low-rank prior aimed to address the challenges of data scarcity for the conditional MAIL framework. Finally, we run a wide range of experiments touching on all three methods of offline/static/online evaluation. Our experiments suggest that our tensor decomposition model with low-rank inductive bias can adapt better than baseline methods such as meta-learning, multi-task learning, or non-linear latent modelling approaches.

In particular, we experiment with our model across various settings with both AI/human partner distributions and up to \(60\) dimensional state spaces with both continuous/discrete actions. We study a collaborative contextual multi-armed bandit task, a continuous-action particle environment, and the card game Hanabi. For these environments, we generate partners by training AI agents with different random seeds. We then do a more comprehensive evaluation on the task of Overcooked~\cite{Carroll2019OnTU,wang2020too}, which consists of human-human gameplay on \(5\) different layouts of the game. We deploy our adaptive policy learned from these human-human demonstrations in a user study with crowd-sourced workers on Prolific, and our study suggests that our models is robust to non-stationary human partners as well.

\section{Related Work}

\noindent \textbf{MARL.} Many advancement have been made in multi-agent RL in training a group of agents to succeed at a task together, typically via centralized learning~\cite{lowe2017ma,foerster2018counterfactual}. In multi-agent imitation learning, techniques have been proposed based on adversarial training~\cite{song2018multi,yu2019multi} or structure learning in the setting of unassigned roles~\cite{le2017coordinated}. In these works, the focus is on training policies for one full set of agents in the environment, whereas our goal is to train a policy to adapt to new partners.

\noindent \textbf{Non-stationarity.} Conditional imitation learning fits under the general framework of non-stationarity in the other agent's behavior~\cite{hernandez2017survey}. 
There are many types of non-stationarity, with one line of work focusing on learning procedures that can take advantage of the other agent’s learning process~\cite{zhang2010multi,foerster2018lola} or latent strategy dynamics~\cite{xie2020learning, wang2021influencing}. 
The non-stationarity in our problem arises from the introduction of new partners, and from the strategy drift of our partners over time. This is related to context-detection~\cite{da2006dealing,hernandez2016bayesian} techniques that identify the partner behavior as switching between one of a finite number of stationary strategies. In contrast, our approach can fit a policy for a new partner with a continuous latent space over strategies, and induces on a low-rank prior. Also related are convention modeling approaches~\cite{shih2021on} which target adaptation to new partners using modular policy networks.

\noindent\textbf{Partner Modeling. } Many works on partner modeling have been successful with predicting a human partner's intentions for robotics~\cite{awais2010human,DraganS12Formalizing,jeon2020shared,Javdani-RSS-15}, motion planning~\cite{nikolaidis2017human}, games~\cite{nguyen2011capir}, driving~\cite{Sadigh2016InformationGA}, and more~\cite{lemaignan2017artificial,kok2020trust,chen2020trust}. In~\cite{nikolaidis2015efficient}, they first run unsupervised clustering to learn partner types, and then run a solver on the MDP taking into account the inferred partner types. These works generally require a strong model of the environment so that after inferring the partner intent, they are able to incorporate the intent into a planner for the ego agent. Our framework differs in that we train only on trajectory datasets, without access to environment reward, dynamics, or a planner. This makes our setting challenging, but more applicable to general tasks for which we do not have knowledge of the environment.

\noindent \textbf{Imitation Learning.} Since we are working with multiple partners, we can draw similarities to hierarchical/conditional imitation learning methods that context switch between a set of low-level policies~\cite{codevilla2018end,le2018hierarchical}. From this perspective, our approach is related to learning a continuous range of low-level policies, and choosing between them based on inference over the partner strategy. The framework of meta-imitation learning~\cite{finn2017one} is also relevant, where they learn to solve a new RL task with just a single expert demonstration. On the other hand, there are works that learn from heterogeneous demonstrations too~\cite{li2017infogail,paleja2020interpretable}. The main difference between our setting and the single agent IL methods are that their test environment has no non-stationarity, and that their demonstrations are directly over the experts, whereas our goal is to predict expert actions given demonstrations of the partner.

\noindent \textbf{Sequence Prediction.} Multi-agent trajectory prediction~\cite{AlahiGRRLS16,gupta2018social,pokle2019deep} and multivariate time-series prediction~\cite{chakraborty1992forecasting} are closely related to the framework of predicting the expert actions given the partner actions. Whereas some works~\cite{AlahiGRRLS16} aim to predict the actions of every agent given the past actions of every agent, our setting aims to predict the expert actions given access to only the partner actions.

\section{Problem Statement}

\begin{figure}[htbp]
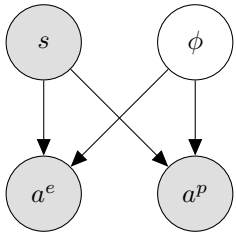

    \begin{minipage}{0.5\linewidth}
        \centering
        \tikz{
         \tikzset{latent/.append style={minimum size=1cm}}
        
         \node[obs] (a_p) {$a^p$}; %
         \node[latent,above=of a_p] (phi_p) {$\phi$}; %
         \node[obs,left=of phi_p] (s) {$s$}; %
         \node[obs,left=of a_p] (a_e) {$a^e$}; %
        
         \edge {s,phi_p} {a_p}
         \edge {s,phi_p} {a_e}
        }
    \end{minipage}
    %
    \begin{minipage}{0.43\linewidth}
        \centering
        \caption{Partner strategies \(\phi\) are sampled from an underlying distribution \(\Phi_\text{ptnr}\). The strategy of a partner affects its actions \(a^p\) taken at a state \(s\). We model the experts as having inferred the partner strategy, so their action \(a^e\) at state \(s\) also depends on \(\phi\).}
        \label{fig:graphicalmodel}
    \end{minipage}

\end{figure}

We consider a two-player Markov Game with states \(S\), and actions \(\A = A \times A\). Agents take actions independently of each other, i.e. \(\pi(\ba|s) = \pi^0(a|s) \pi^1(a|s) \). Without loss of generality, the ego agent policy (that we control) is \(\pi^0\) and the partner policy is \(\pi^1\).

One standard framework in multi-agent imitation learning is to train a single pair of policies \((\pi^0, \pi^1)\) to mimic trajectories from a pair of experts~\cite{song2018multi}. This framework, however, does not consider generalization to new partners. We are interested in the adaptive setting -- can we train an agent to play well against a new partner, whose actions come from a different policy than that of a partner from the training set?

We assign each partner an ID \(y\) and strategy \(\phi\) as a sample drawn from some fixed distribution \(\Phi_{\text{ptnr}}\). We assume that the agents generating the training dataset are stationary, having converged to the stochastic policy \(\pi_\phi(a|s, \phi)\). Since we are in the imitation learning setting, for each partner \(y\) we have joint trajectories \(D^y = \{ (s_i, a^{e}_i, a^{p}_i) \}_{i=1}^k \), where \(a^{e}\)'s are expert action, and \(a^{p}\)'s are partner actions. We assume that the expert has inferred the partner's strategy \(\phi\) through prior repeated interactions, and acts based on a stochastic policy \(\pi_{\text{expert}}(a|s, \phi)\). Hence, even though the agents take simultaneous actions, we model a dependency from \(\phi\) to \(a^e\).

At training time, we have access to batches of joint trajectories \(D^{y_1}, \ldots, D^{y_m}\), corresponding to supervision for the training partners with underlying strategies \(\phi_1, \ldots, \phi_m \sim \Phi_{\text{ptnr}}\). 
At test time, we interact with a new partner \(\testy\) and adapt our ego agent to best coordinate with the new partner.

\subsection{Evaluation Metrics}

As alluded to earlier, there are many evaluation metrics we can consider when interacting with the test partner.

\subsubsection{Offline Evaluation}
One option is a purely offline evaluation without the environment, in which case we require joint trajectories in the test set for evaluation. We split the joint trajectories from the test set \(D^{y_0}\) into partial trajectories consisting of the partner actions \(\tilde{D}^{\testy}\) and the expert actions \(\hat{D}^{\testy}\). Under this setup, adapting to a new partner \(\testy\) with strategy \(\phi_0\) corresponds to generating a good ego-agent policy given observations of the partner's actions \(\tilde{D}^{\testy}\). We write our objective as minimizing the KL-divergence between the (unknown) expert policy and the generated ego-agent policy, denoted as \( \pi(a | s, \tilde{D}^{\testy}) \).

\begin{align}
    & \inf_\pi \,\, \sum_{s_i \in \tilde{D}^{\testy}} KL(\pi_{\text{expert}}(a | s_i, \phi_0) \, \vert \vert \,  \pi(a | s_i, \tilde{D}^{\testy})) \notag \\
    = & \sup_\pi \,\, \sum_{s_i \in \tilde{D}^{\testy}} \EX_{\pi_{\text{expert}}(a | s_i, \phi_0)}[ \log \pi(a | s_i, \tilde{D}^{\testy}) ] \label{eq:MLE}
\end{align}

As the entropy of the expert policy is a constant, this objective corresponds to maximizing the log-likelihood of the actions of the expert policy. Although we don't have access to the expert policy, we estimate the objective by treating the expert trajectories \(\hat{D}^{\testy}\) as samples from the expert policy.

\subsubsection{Static Evaluation}
If we have access to the environment dynamics and rewards at test time, we can pair our ego agent with new partner agents and measure the environment reward. Either the partner agent policies are provided at test time, or we behavioral-clone their policies from a test dataset.

In both offline and static evaluation, the partner agent is stationary, governed by some fixed strategy. Based on the formulation in Figure~\ref{fig:graphicalmodel}, we can then write our desired policy as a mixture of how an expert would react to each partner type, weighted by the probability of each partner type given observations \(\tilde{D}^{\testy}\).
\begin{align}
    \pi(a | s, \tilde{D}^{\testy}) 
    = & \int_{\phi} \pi_{\text{expert}}(a | s, \tilde{D}^{\testy}, \phi) p(\phi | s, \tilde{D}^{\testy}) d \phi \notag\\
    = & \int_{\phi} \pi_{\text{expert}}(a | s, \phi) p(\phi | \tilde{D}^{\testy}) d \phi \label{eqn:policypartnertype}
\end{align}

\subsubsection{Online Evaluation}
Lastly, we can relax our assumptions and measure the environment reward attained when pitting our adaptive ego agent against non-stationary partners at test time. The test partner policies may drift over time, and may not correspond to a fixed strategy \(\phi\). This requires us to continuously finetune our ego agent policy as new interaction data is collected. Notably, for this online evaluation method, we can pair our adaptive policy against crowd-sourced human players, whose behavior will be naturally influenced by the actions that our ego agent makes.

The highlight of the conditional multi-agent imitation learning framework is that training only requires a static dataset of joint trajectories from diverse partner/expert pairs, making it generalizable to many tasks in the real world. As we have seen, the general aim is to learn a conditional expert policy \(\pi_{\text{expert}}(a | s, \phi)\) over a latent space of strategies \(\phi\), and to infer \(\phi\) using observations of a new partner's actions at test time. By continuously updating \(\phi\) based on newly collected trajectories, the ego agent can adapt even to non-stationary partners, all without training on the environment rewards or dynamics.

\section{Learning a Low Rank Policy Space}

Learning a conditional expert policy \(\pi_{\text{expert}}(a | s, \phi)\) along with a model \(p(\phi | \tilde{D}^{\testy})\) for inferring partner strategy is challenging. In addition, learning a non-linear latent space over strategies can struggle with overfitting when data is limited.

Instead we propose to learn a low-rank latent space that is both scalable to high dimensional environments and suitable for low data settings. Our method aims to impose a more structured prior on the latent space that may reduce the expressiveness of the model, but in return cut back on the model's reliance on large amounts of data.

Given the scalability and data-efficiency benefits of imposing a low-rank prior, the main question is then -- how restrictive is a low-rank prior on the model's ability to learn a good latent space over strategies for conditional MAIL? Based on existing evidence in the literature, we argue that multi-agent tasks have two significant sources of structure, which suggest that the space of reasonable policies may indeed lie on a very low-dimensional subspace.

\begin{itemize}[]
    \item \textbf{Task constraints} -- the reward and dynamics of the Markov Game can be viewed as imposing soft constraints, ruling out actions that are clearly suboptimal. These soft constraints, in the form of Q-functions of the environment, can filter out the majority of possible policies, in particular those that often take low-value actions. Recent work has shown that the Q-functions of RL environments have surprisingly low rank~\cite{gorodetsky2018high, Yang2020Harnessing}. 
    
    \item \textbf{Partner Similarities} -- the partners from the distribution \(\Phi_\text{ptnr}\) may only take on a small subspace of possible strategies, both when considering human or AI partners. For example, exploiting common similarities between humans is what drives the success of collaborative filtering methods~\cite{schafer2007collaborative,he2017neural,yin2021ttrec}. Beyond human partners, it has also been observed that neural network agents trained via stochastic gradient descent span a surprisingly low-dimensional subspace \cite{li2018visualizing,gur2018gradient}.
\end{itemize}

In summary, the underlying Markov Game already rules out a large swath of undesirable policies, and the partner sampling distribution further restricts the space of possible partners. With this in mind, we explore the use of a low rank model to capture the existing structure in the latent space of strategies. 

Our proposed model makes use of low-rank tensor decomposition. In particular, we have three dimensions of interest in this low-rank structure: strategy \(\phi\), states \(s\), and actions \(a\). 
Unlike its matrix counterpart, the low-rank tensor decomposition does not have a unique formulation. Popular methodologies include the CANDECOMP/PARAFAC decomposition~\cite{carroll1970analysis,Harshman1970FoundationsOT}, the Tucker decomposition~\cite{tucker1966some} and the Tensor Train decomposition~\cite{cichocki2016low}. In this work, we focus on the Tensor Train decomposition due to its scalability properties~\cite{oseledets2011tensor,bigoni2016spectral,yin2021ttrec}.

\subsection{Tensor Train Decomposition}

The Tensor Train decomposition is a compact and scalable representation of high-dimensional functions. A Tensor Train of rank \(r\) can represent an \(n\)-dimensional function (each dimension taking on \(I\) values) with \(O(nIr^2)\) parameters, whereas a na\"ive representation requires \(O(I^n)\) parameters. Apart from scaling better than other low-rank tensor decomposition alternatives, the Tensor Train also comes equipped with rank-reducing approximation algorithms~\cite{oseledets2011tensor,cichocki2016low}.

The discrete form of a Tensor Train represents a high-dimensional function \(g : I_1 \times \ldots \times I_n \rightarrow \mathbb{R}^k\) by keeping \(n\) ``cores'' (i.e. \(3\)-dimensional tensors) of shape \(A_i = r_{i-1} \times I_i \times r_i\). The values \([r_0, \ldots, r_n]\) are the rank of the tensor, where each \(r_i\) is an integer, with \(r_0 = 1\) and \(r_n = k\). For convenience, we will refer to a Tensor Train with \(r_1 = \ldots = r_{n-1} = r\) simply as having rank \(r\). The values \(I_1, \ldots, I_n\) are the modes of the Tensor Train, indicating the number of values the input to each dimension can take on. To evaluate a tensor train on an input \(x_1 \ldots x_n\), we simply index into the cores at each dimension (resulting in a series of matrices), and perform a series of matrix multiplication.
\begin{align*}
    g(x_1, \ldots, x_n) = A_1[:,x_1,:] \times \ldots \times A_n[:,x_n,:]
\end{align*}
The intermediate matrices have shape \(1 \times r\) and \(r \times k\) at the endpoints, and \(r \times r\) in between, so the series of matrix multiplications will produce an output value in \(\mathbb{R}^k\) as desired. 

We would like to directly use the Tensor Train \(g\) as our policy network, by having two cores: a partner-strategy core and a state core. Then we set the output dimension \(k\) to be equal to the size of the action space \(|A|\). But, we face two issues. First, the mode of the state core (e.g. the total number of states in the state space) may be too large. Second, we only observe the partner identities \(y\), and not the partner strategies \(\phi\), so we need to interpret tensor core for the partner strategies differently. We address these issues next.

\subsection{Functional Tensor Train}

Although the Tensor Train format scales favorable with respect to the dimensionality, the modes of each dimension \(I_1, \ldots, I_n\) can still be too large (and infinite for continuous inputs) for some practical applications. For example, one of our dimensions of interest is the state \(s\) of a Markov Game, which would correspond to a mode equal to the \emph{total number of possible states}. For most environments, storing a Tensor Train in this format is unmanageable.

An appealing solution is that of the functional Tensor Train~\cite{bigoni2016spectral, gorodetsky2019continuous}. The key insight is that, instead of representing cores with tensors of size \(r_{i-1} \times I_i \times r_i\), we can generalize them to matrix-valued functions \(g_i : I_i \rightarrow \mathbb{R}^{r_{i-1} \times r_i}\). Indexing into a core to retrieve an \(r_{i-1} \times r_i\) matrix now becomes evaluating the matrix-valued function in each dimension:
\begin{align*}
    g(x_1, \ldots, x_n) = g_1(x_1) \times \ldots \times g_n(x_n)
\end{align*}
Previous works limit themselves to piece-wise polynomials in these matrix-valued functions~\cite{gorodetsky2019continuous}, in order to support operations such as integration and cross-approximation. In this work, we instead propose to parameterize each matrix-valued function with a neural network, forgoing support for these operations in order to maximizing flexibility. Using the functional variant of Tensor Trains, we can handle dimensions with extremely large modes or even continuous inputs.

\begin{figure}
    \begin{minipage}{0.45\linewidth}
        \centering
        \includegraphics[width=\linewidth]{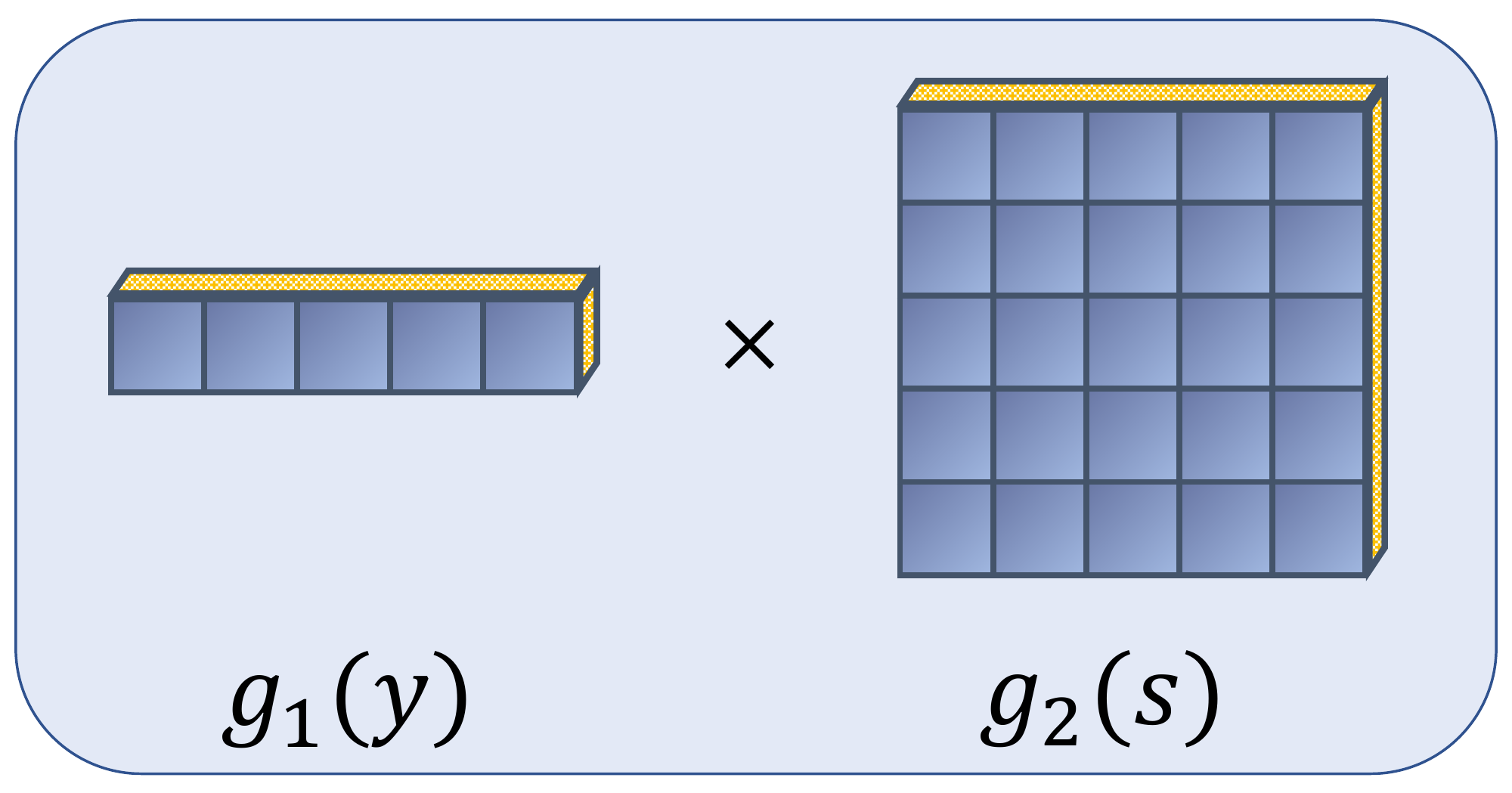}
    \end{minipage}
    \hfill
    \begin{minipage}{0.5\linewidth}
        \centering
        \caption{
        Given an observation \(s\) from a partner with ID \(y\), the functional Tensor Train evaluates a matrix-valued outputs \(g_1(y), g_2(s)\). Then the Tensor Train performs matrix multiplication, and the resulting vector is the action logits for the policy.}
        \label{fig:tt}
    \end{minipage}
\end{figure}

\subsection{Conditional Policies}

We next describe how to use the functional Tensor Train to fit policies of different partners strategies. We want to represent the latent space over partner strategies using the first core of our functional Tensor Train.
For any given partner, we only observe its identity \(y\) and not its strategy \(\phi\). Therefore, we let the partner strategy core of our Tensor Train take in the (one-hot vector) identity \(y\) as input, and use the \(1 \times r\) matrix-valued output as the partner strategy \(\phi\). This is not restrictive, since the neural network mapping \(\phi = g_1(y)\) is flexible.

The setup of our architecture is as follows (Figure~\ref{fig:tt}). We have the neural network (strategy core) that maps the one-hot partner ID to a \(1 \times r\) vector \(\phi\) -- the latent representation of the partner strategy. We have another neural network (state core) that maps the state observation to a \(r \times |A|\) matrix. The state encodes the role of the agent (ego or partner), so the same policy network is used for both the ego and the partner. The result of the matrix multiplication is used as the logits for the policies, and this enforces a low-rank prior in the log-probabilities of the actions.
\begin{align}
    \log \pi(A|s) = \frac{1}{Z} g_1(y) \times g_2(s) \label{eq:logits}
\end{align}
At train time we learn the parameters of \(g_1, g_2\) via Maximum Likelihood Estimation (MLE) on both the ego and the partner actions. The objective is Equation~\ref{eq:MLE} for both the ego and partner agent, plugging in their policies as defined by Equation~\ref{eq:logits}. 

At test time, given a new test partner, we set its partner ID to a random vector \(y_0\). This gives us a random initialization of the strategy vector \(\phi = g_1(y_0)\). Then as we interact with the test partner, we can fine-tune the strategy core \(g_1\) via MLE on the newly collected test partner's actions (e.g. Equation~\ref{eq:MLE} on the partner actions).

The structure of the Tensor Train makes it easy to finetune to new partners at test time. Given a new partner with ID \(y_0\), we fine-tune only the partner-strategy core of the Tensor Train, keeping the parameters of the state core fixed. 
In essence, the learning phase during training is fitting a low-rank subspace over policies for different partners, and the adaptation phase at test time is doing inference over the partner strategy and interpolating the output actions in that subspace.
Moreover, we do not need to specify the number of testing partners in advance, since we can randomly initialize the ID vector \(y_0\) and update the mapping \(g_1(y_0)\). Finally, our method avoids overfitting by enforcing a low-rank inductive bias.

\section{Experiments}

We run a variety of experiments to evaluate our approach for adapting to new partners in multi-agent games, spanning discrete/continuous actions, full/partial observability, and AI/human partners.
For the first set of experiments, we generate partner distributions by training AI partners with different random seeds. We study a collaborative multi-armed bandit task, a particle environment with continuous actions, and the game of Hanabi using \(4\) players. For the second set of experiments, we explore the distribution over human partners instead. We focus on the game of Overcooked~\cite{Carroll2019OnTU, wang2020too} by training and performing static evaluation on human-human demonstrations collected from~\cite{Carroll2019OnTU}. Then, we follow it up with online evaluation against real human partners, by deploying our adaptive agents in a user study to play Overcooked with crowd-sourced humans. Our experiments show the applicability of our framework to a broad range of different settings. We describe the tasks in detail in their corresponding sections.

We compare our approach ({\bf lrp}: low-rank partners) with four baselines: a meta-learning ({\bf maml})~\cite{finn2017model}, a multi-task learning ({\bf mt})~\cite{caruana1997multitask}, and a modular policy ({\bf mod})~\cite{devin2017learning,shih2021on} approach, and a non-linear latent space approach ({\bf lt}). We implement these methods using the Garage toolkit~\cite{garage}.

The meta-learning approach optimizes for performance after taking inner gradient descent steps on a new sampled task. In our experiments we take one inner gradient step, which has been shown~\cite{nichol2018first} to also give good performance.
The multi-task approach aims to share representations across policies for different partners, by appending the partner ID to the state observations and predicting with a single network. 
The modular policy approach aims to separate task and partner representations with modular policy networks, in order to transfer the task representation to a new partner.
Finally, the non-linear latent space method concatenates learned partner embeddings with state observations, and maps this joint input directly to the action space. 
These baseline methods can work well when the number of training tasks is large, but unlike our method, they do not have a strong (low-rank) inductive bias. This can make it difficult for them to generalize well when the number of training tasks is small.

\subsection{Collaborative Bandits}%
In our collaborative bandit environment, two players simultaneously pick an action, scoring a point if and only if they chose the same action and the score for that action is \(1\). We design our bandit environment to have \(1000\) states, with an action space of \(10\). At each state, roughly \(30\%\) of the actions give a score of \(1\), and the rest give a score of \(0\). In other words, the coordination challenge is in breaking the tie between the equally optimal joint actions.

We generate partners by training AI agents using different seeds. The AI agents are trained to output, at a given state, any one of the actions with a score of \(1\). We train a set of \(16\) training partners and \(4\) testing partners using the same RL algorithm but each with a different seed. The random seed affects the sampled states of the environment, the network initialization for each agent, and the stochasticity in the optimization process.

First, we can experimentally check that these partner policies generated by self-play with different seeds are indeed low-rank. We tabularize the policy distribution of all the partners as a tensor T of size [1000,10,16] (states, actions, \#partners), and fit a low-rank Tensor Train to T (Table~\ref{tab:mab}). The experiment suggests that the tabularized policy tensor has a low rank (of \(4\)), which is in line with phenomena observed by~\cite{gur2018gradient,li2018visualizing}.

\begin{table}[htbp]
    \caption{Fitting Tensor Trains with various ranks to partner policies trained with different random seeds on the bandit env. The model fits the data well from a rank of \(4\) onwards.}
    \label{tab:mab}
    \centering
    \begin{tabular}{c|ccccccc}
        \toprule
         {\bfseries rank} & 1 & 2 & 3 & 4 & 5 & 6 & 7 \\
         \midrule
         {\bfseries log-loss} & 38.36 & 38.28 & 37.90 & 22.09 & 22.03 & 22.11 & 22.21 \\
         \bottomrule
    \end{tabular}
\end{table}

\begin{figure}
    \centering
    \includegraphics[width=\linewidth]{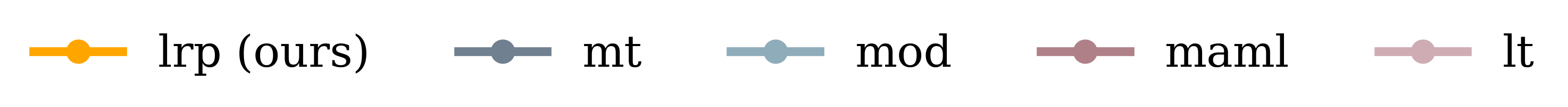}

    \begin{minipage}{\myiconwidth}
        \centering
        \includegraphics[height=0.45\linewidth]{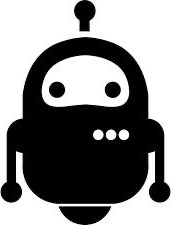}\\
        \includegraphics[height=0.45\linewidth]{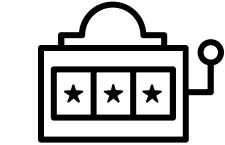}
    \end{minipage}
    \begin{minipage}{\mygraphwidth}
        \centering
        \includegraphics[width=\linewidth]{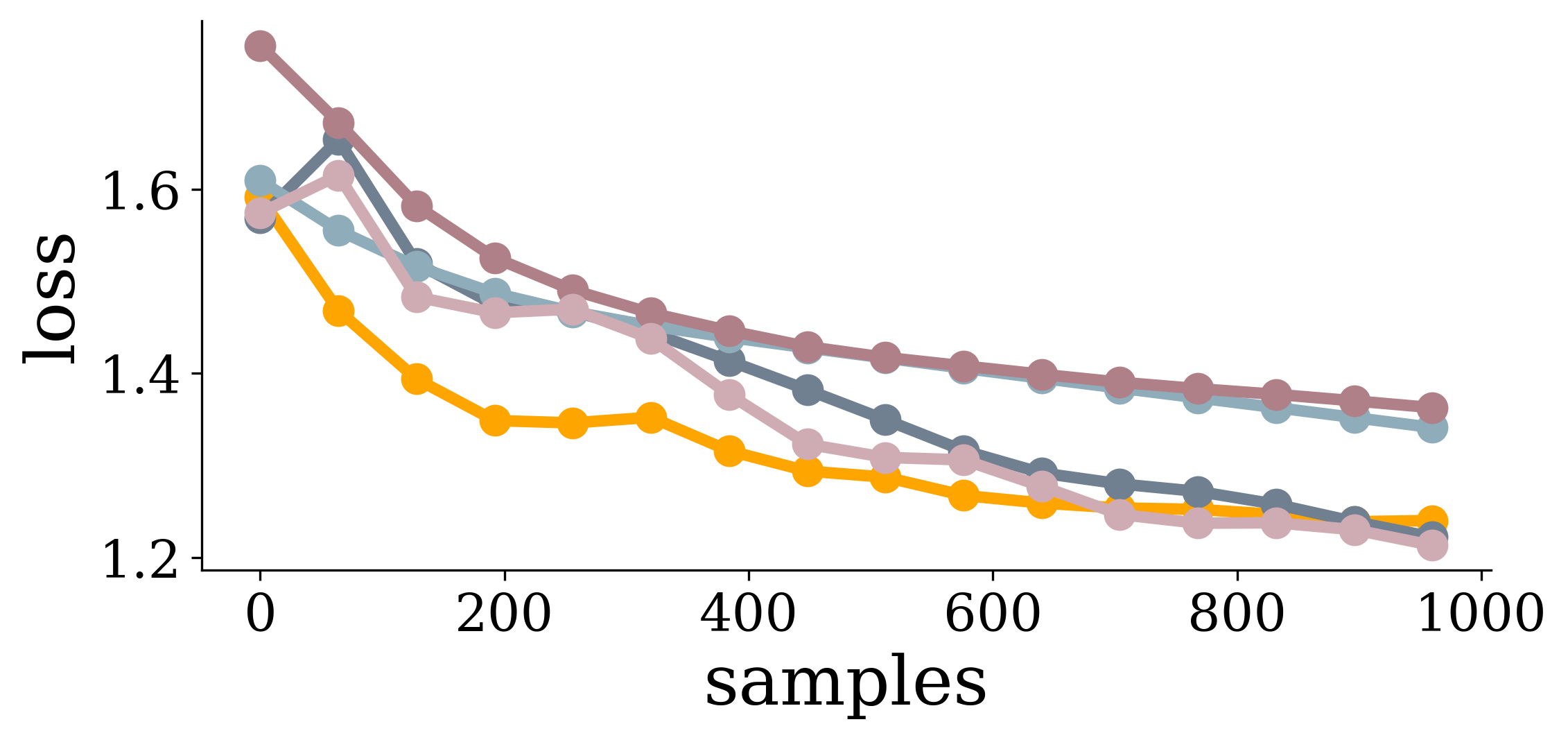}
    \end{minipage}

    \caption{Collaborative bandit task. The partner policies are generated by self-play with different random seeds. We use \(16\) training and \(4\) testing partners. We set the rank of the Tensor Train to \(4\). We compute the loss of adapting to each new testing partner, and plot the average loss over all the testing partners.}
    \label{fig:mab}
\end{figure}

In Figure~\ref{fig:mab}, we plot the performance of different techniques for adapting to a new partner. The loss is the negative log-likelihood of the held-out expert actions for the test partners, which for the bandit task is equivalent to the partner actions. We see that our low-rank partner approach adapts best to new partners at test time from this partner distribution arising from self-play.
This suggests that our low-rank partner approach is indeed capturing the low-rank subspace of the data-generating process, as revealed by Table~\ref{tab:mab}, and correctly inferring the strategy of the new partner.
The underlying Tensor Train is able to leverage the correlation between actions at different states (due to the low-rank structure), and adapt quickly instead of learning the action distributions at each state separately.

\subsection{Particle Environment}
\label{sec:particle}

\begin{figure}
    \begin{minipage}{0.45\linewidth}
        \centering
    \includegraphics[width=\linewidth]{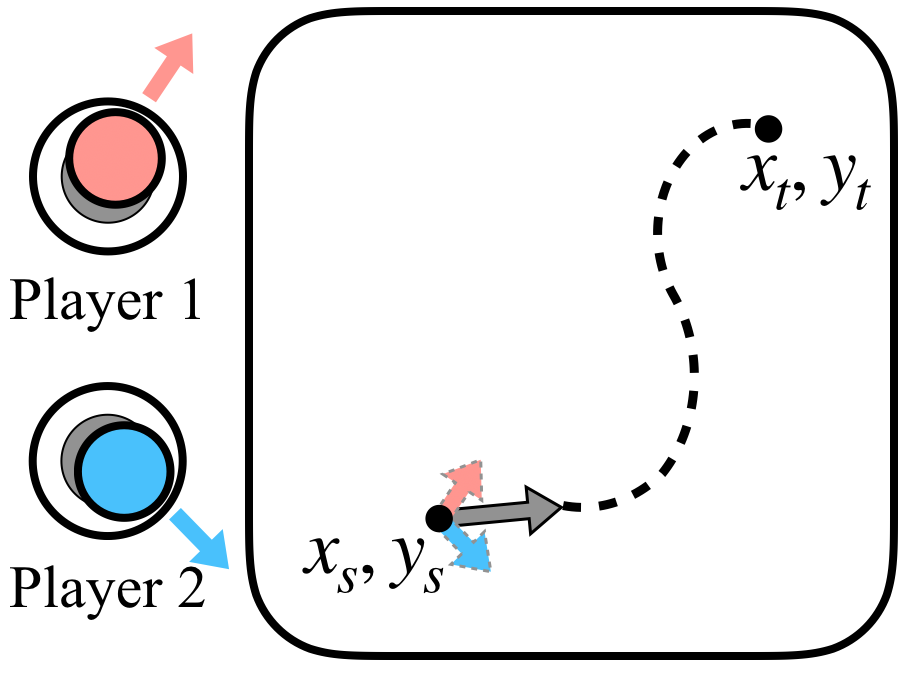}
    \end{minipage}
    \hfill
    \begin{minipage}{0.5\linewidth}
        \centering
        \caption{We visualize the \(2\)-player collaborative particle environment. The two players work together to move a particle on a 2D plane from from position \(x_s,y_s\) to \(x_t,y_t\). The players input 2D action vectors simultaneously, and the position of the particle moves according to the sum of the action vectors of the two players at each timestep.}
        \label{fig:particle_diagram}
    \end{minipage}
\end{figure}

To test our model on continuous actions, we consider a 2D collaborative particle environment where the two players have to move a particle from a start location \(x_s, y_s\) to a target location \(x_t, y_t\) (Figure~\ref{fig:particle_diagram}). Each player takes a 2D continuous action representing the velocity of the particle. The task is collaborative -- at each timestep the particle moves according to the sum of the actions (i.e. velocities) given by the two players, with reward being the negative distance to the target. We follow a similar setup as earlier, training policies to play against \(16\) train partners obtained via self-play, and adapt to \(4\) test partners, without any access to environment reward. As an upper bound, the self-play agents are able to achieve a reward of \(313 \pm 7\) when playing with themselves. In Table~\ref{tab:particle} we report the environment reward attained by the adaptive agents before and after adapting to a new partner using a trajectory of \(200\) timesteps. We see that the low-rank partner approach achieves the best reward of the \(5\) compared methods.

\begin{table}[h]
    \caption{Reward in particle env before/after adapting to test partners.}
    \label{tab:particle}
    \centering
    \begin{tabular}{c|ccccc}
         \toprule
         &
         {\bfseries lrp} & 
         {\bfseries mt} &
         {\bfseries mod} &
         {\bfseries maml} &
         {\bfseries lt} \\

         \midrule
         before & \({\bm{297}} \pm 4\) & \(274 \pm 4\) & \(290 \pm 5\) & \(272 \pm 11\) & \(289 \pm 4\) \\
         after & \({\bm{303}} \pm 4\) & \(300 \pm 4\) & \(300 \pm 4\) & \(275 \pm 7\) & \(294 \pm 4\) \\
         \bottomrule
    \end{tabular}
\end{table}

\subsection{Hanabi}

The Hanabi environment differs from the other environments in that it has partial observability, and it involves more than two agents. Nonetheless, the framework of conditional multi-agent imitation learning can be applied to these settings as well. In particular, we can handle the ego-agent as usual, and treat the other \(n-1\) agents as one joint partner agent. Of course, more careful handling of the interaction between the \(n-1\) players is possible, which we leave for future work.

In Hanabi~\cite{bard2020hanabi}, a group of agents (\(4\) in our case) work together to play cards from their hands in a specific order, almost akin to multi-agent Solitaire. The catch is that players can see everyone else's cards except their own. Since their is no communication, the players must develop conventions with each other, hence adapting to a new conventions is critical to success with a new partner.

\begin{figure}
    \centering
    \includegraphics[width=\linewidth]{figs/legend.png}
    \includegraphics[width=0.9\linewidth]{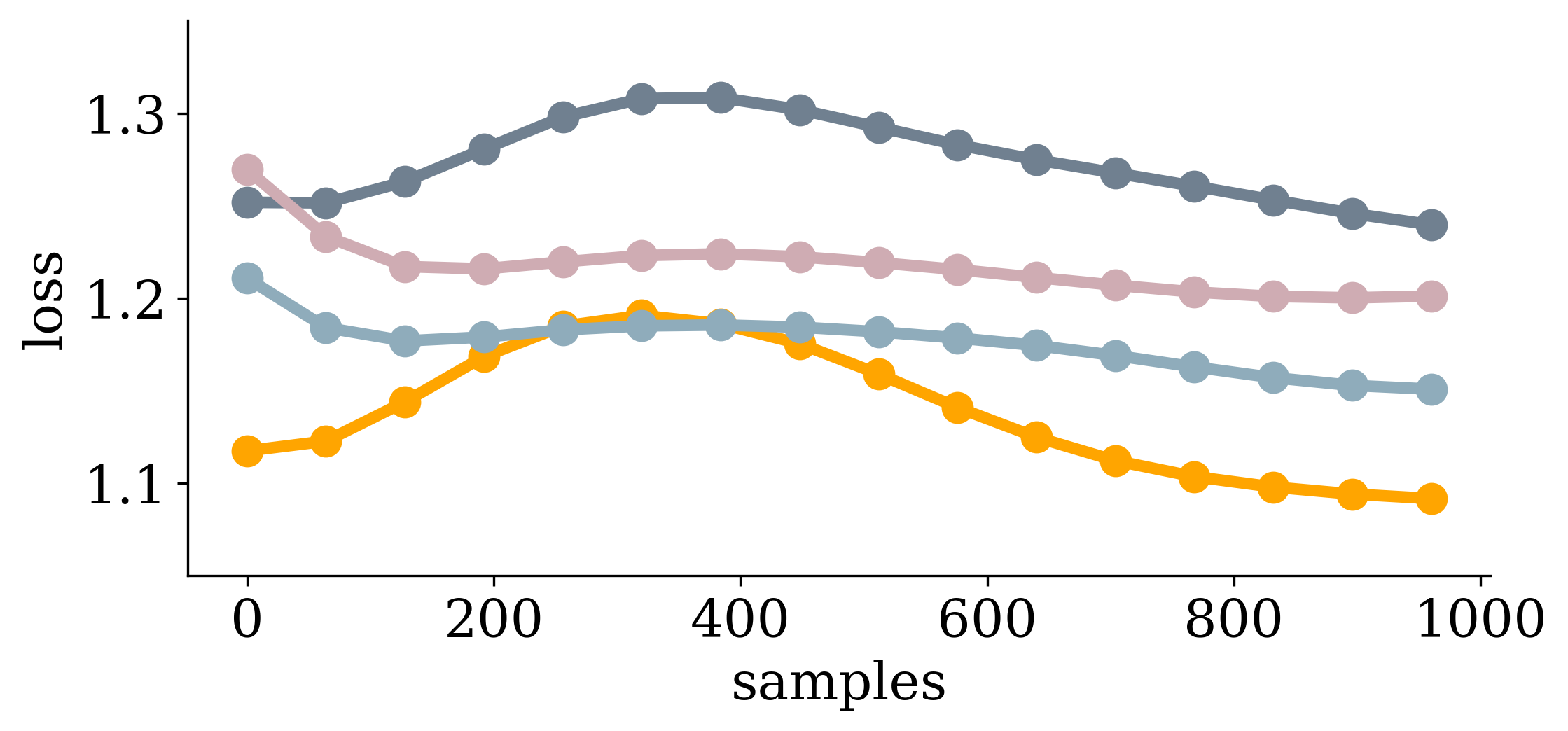}
    \caption{Results from the game of Hanabi with \(4\) players, where our ego agent is adapting to a new set of other \(3\) partners each time. Hanabi is a collaborative turn-based game with partial observability, as every player can only see other player's cards but not their own. In our experiments, our low-rank model showed better adaptation performance after \(1000\) timesteps of demonstrations from a new partner.}
    \label{fig:hanabi}
\end{figure}

We study a small version of the game with \(1\) color, \(5\) ranks, \(4\) players, and hand sizes of \(2\). As before, we generate training and testing partner/expert pairs using self-play with different random seeds. In Figure~\ref{fig:hanabi} we plot the loss (negative log-likelihood) of the policies as they adapt to a new partner, with the \(x\)-axis denoting the size of the demonstration used for adaptation. Of the compared methods, our low-rank approach exhibits the best adaptation performance after \(1000\) samples.

\subsection{Overcooked}%

\begin{figure}
    \centering
    \includegraphics[width=\linewidth]{figs/legend.png}

    \begin{minipage}{\myiconwidth}
        \centering
        \includegraphics[height=\myiconsubheight]{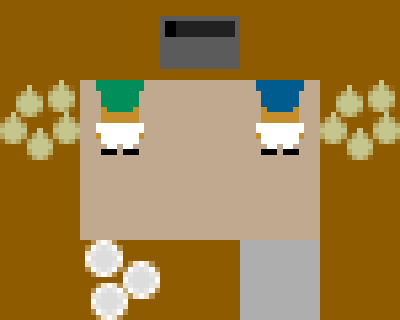}
        \small{Cramped Room}
    \end{minipage}
    \begin{minipage}{\mygraphwidth}
        \centering
        \includegraphics[width=\linewidth]{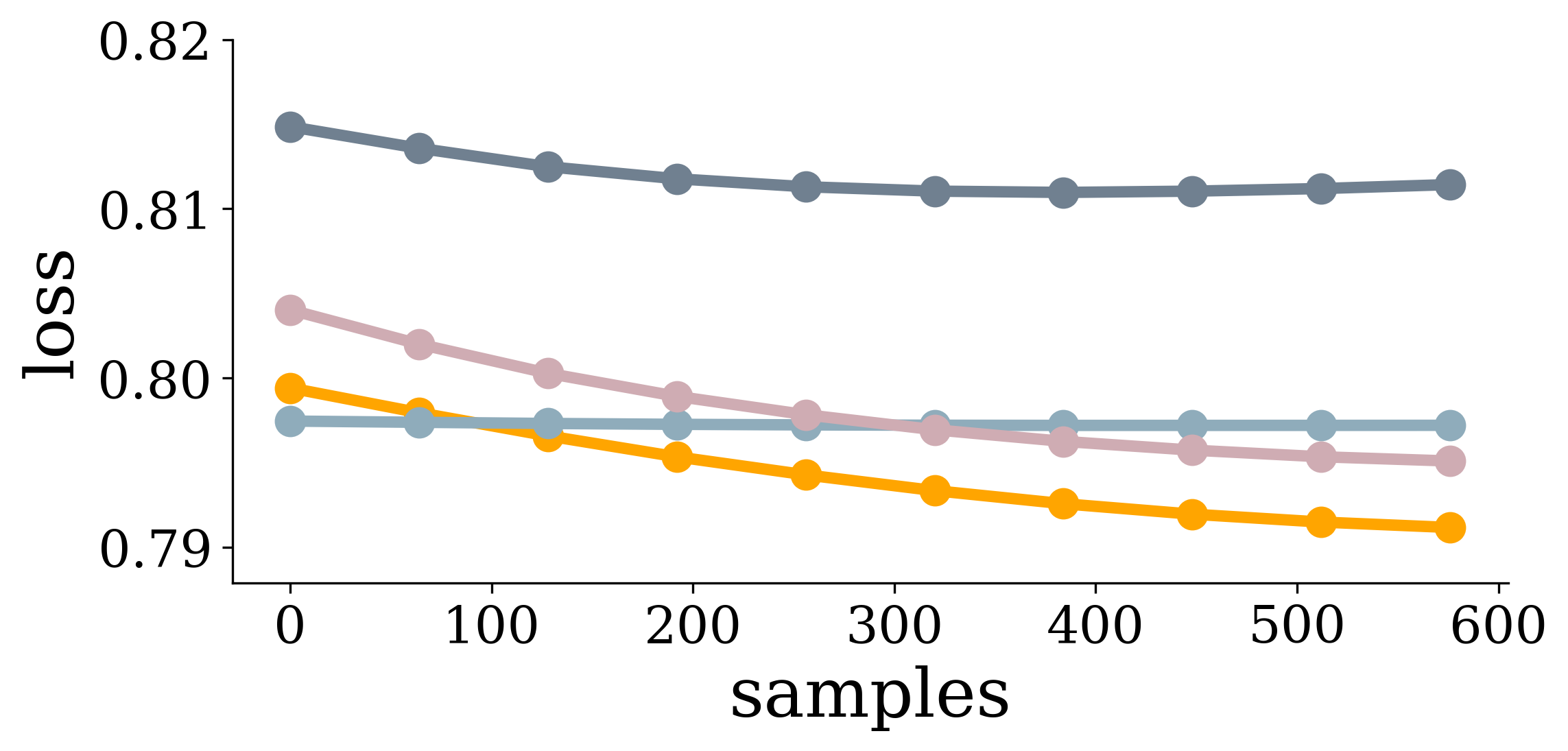}
    \end{minipage}
    \hfill
    \begin{minipage}{\myiconwidth}
        \centering
        \includegraphics[height=\myiconsubheight]{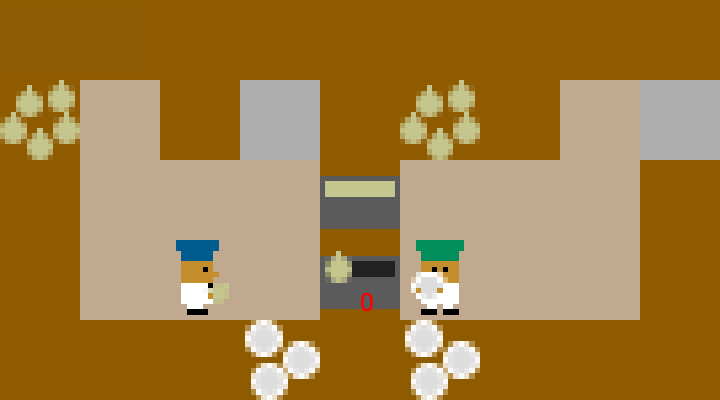}
        \small{Asymmetric Advantages}
    \end{minipage}
    \begin{minipage}{\mygraphwidth}
        \centering
        \includegraphics[width=\linewidth]{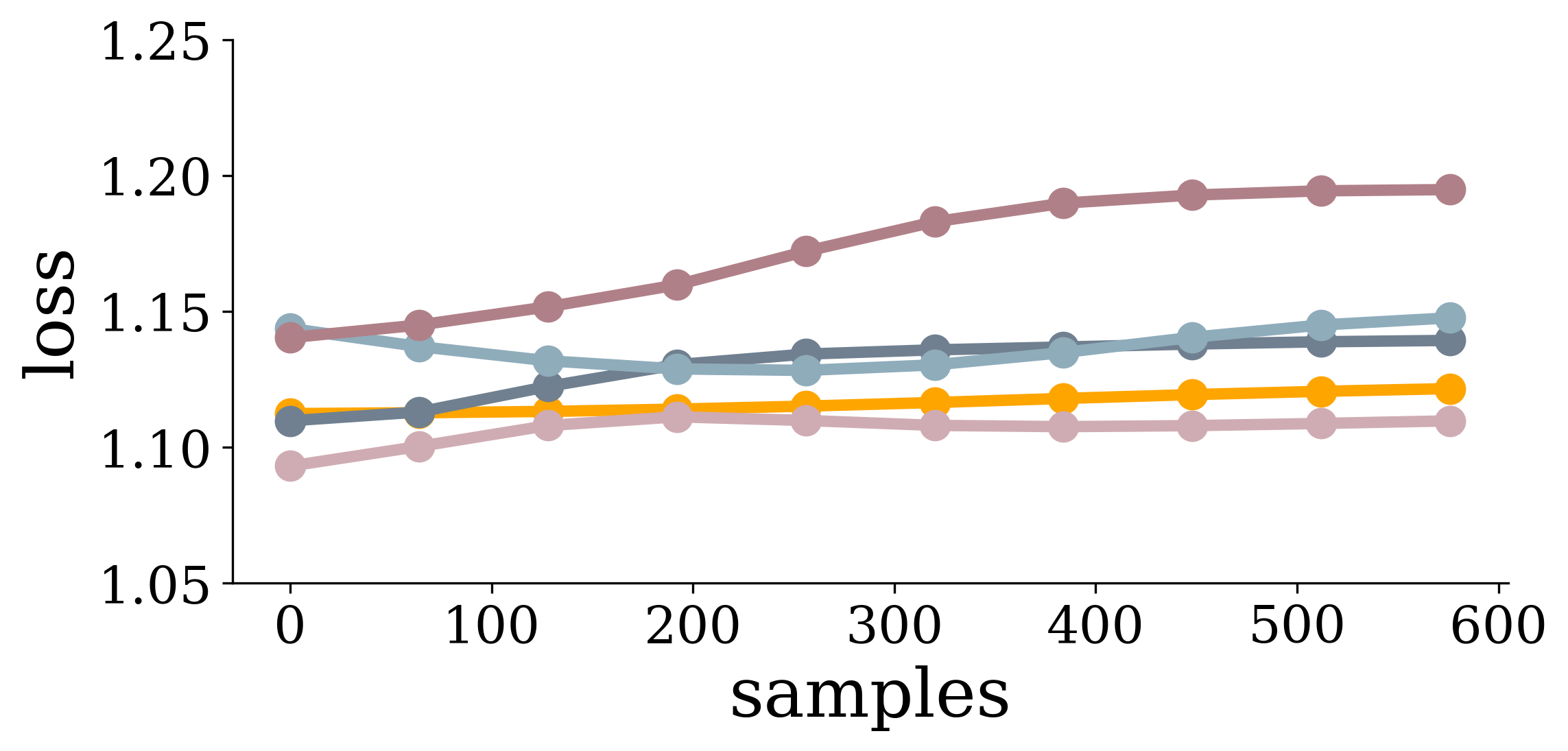}
    \end{minipage}
    \hspace{0.05\linewidth}

    \begin{minipage}{\myiconwidth}
        \centering
        \includegraphics[height=\myiconsubheight]{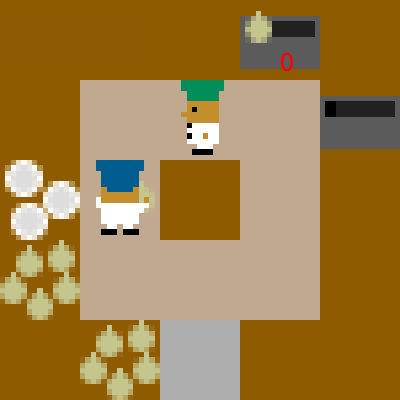}
        \small{Coordination Ring}
    \end{minipage}
    \begin{minipage}{\mygraphwidth}
        \centering
        \includegraphics[width=\linewidth]{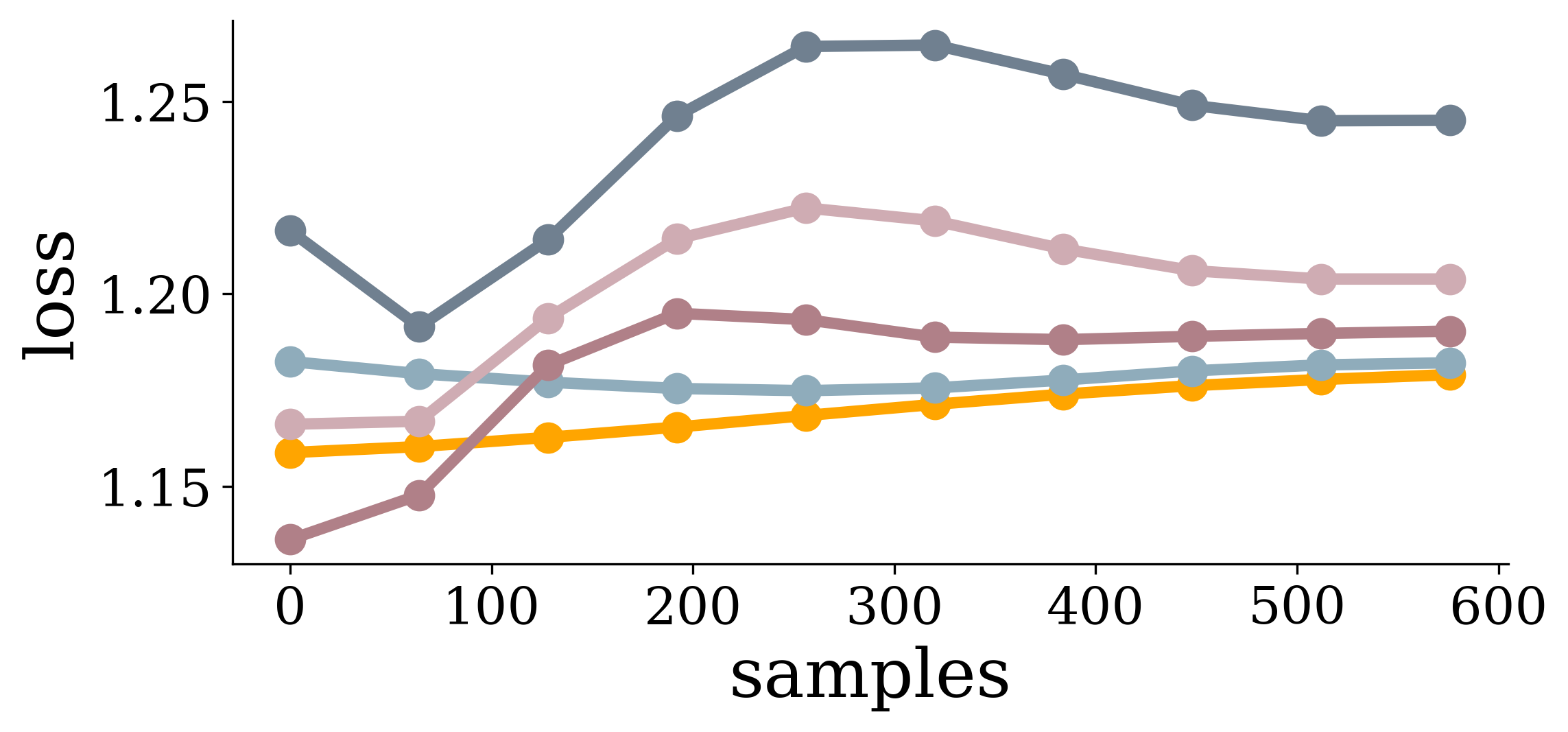}
    \end{minipage}
    \hfill
    \begin{minipage}{\myiconwidth}
        \centering
        \includegraphics[height=\myiconsubheight]{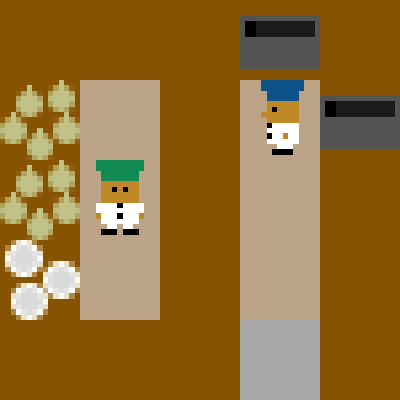}\\
        \small{Forced Coordination}
    \end{minipage}
    \begin{minipage}{\mygraphwidth}
        \centering
        \includegraphics[width=\linewidth]{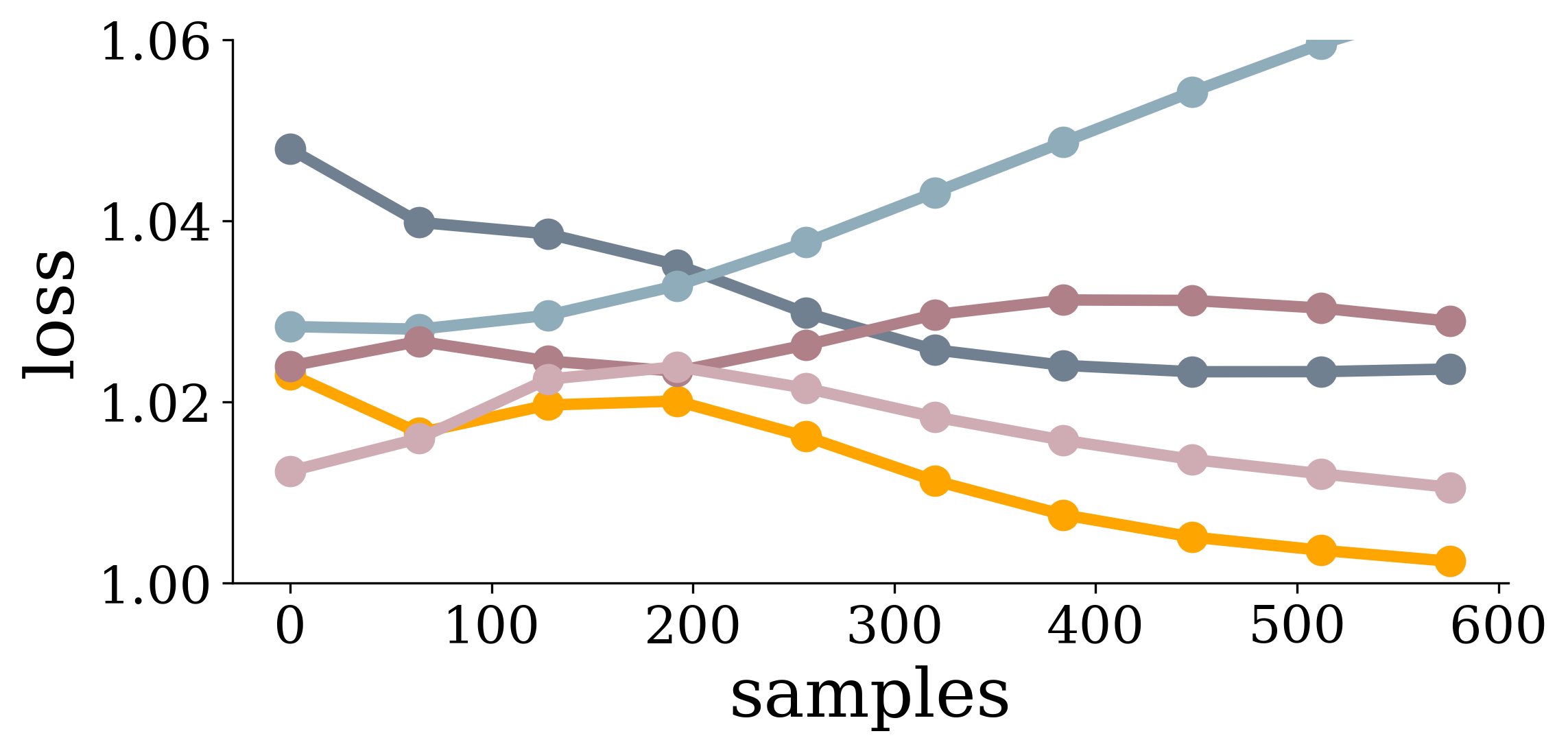}
    \end{minipage}
    \hspace{0.05\linewidth}
    
    \begin{minipage}{\myiconwidth}
        \centering
        \includegraphics[height=\myiconsubheight]{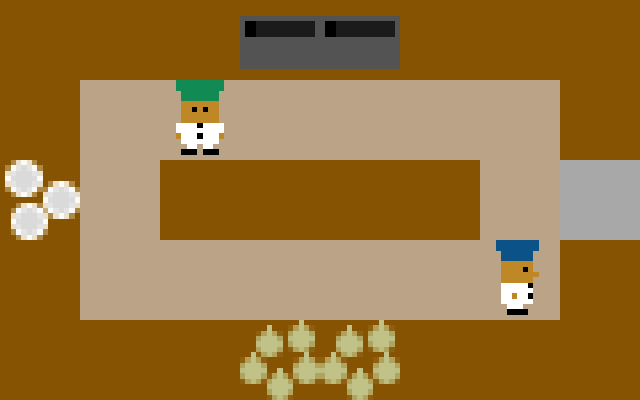}
        \small{Counter Circuit}
    \end{minipage}
    \begin{minipage}{\mygraphwidth}
        \centering
        \includegraphics[width=\linewidth]{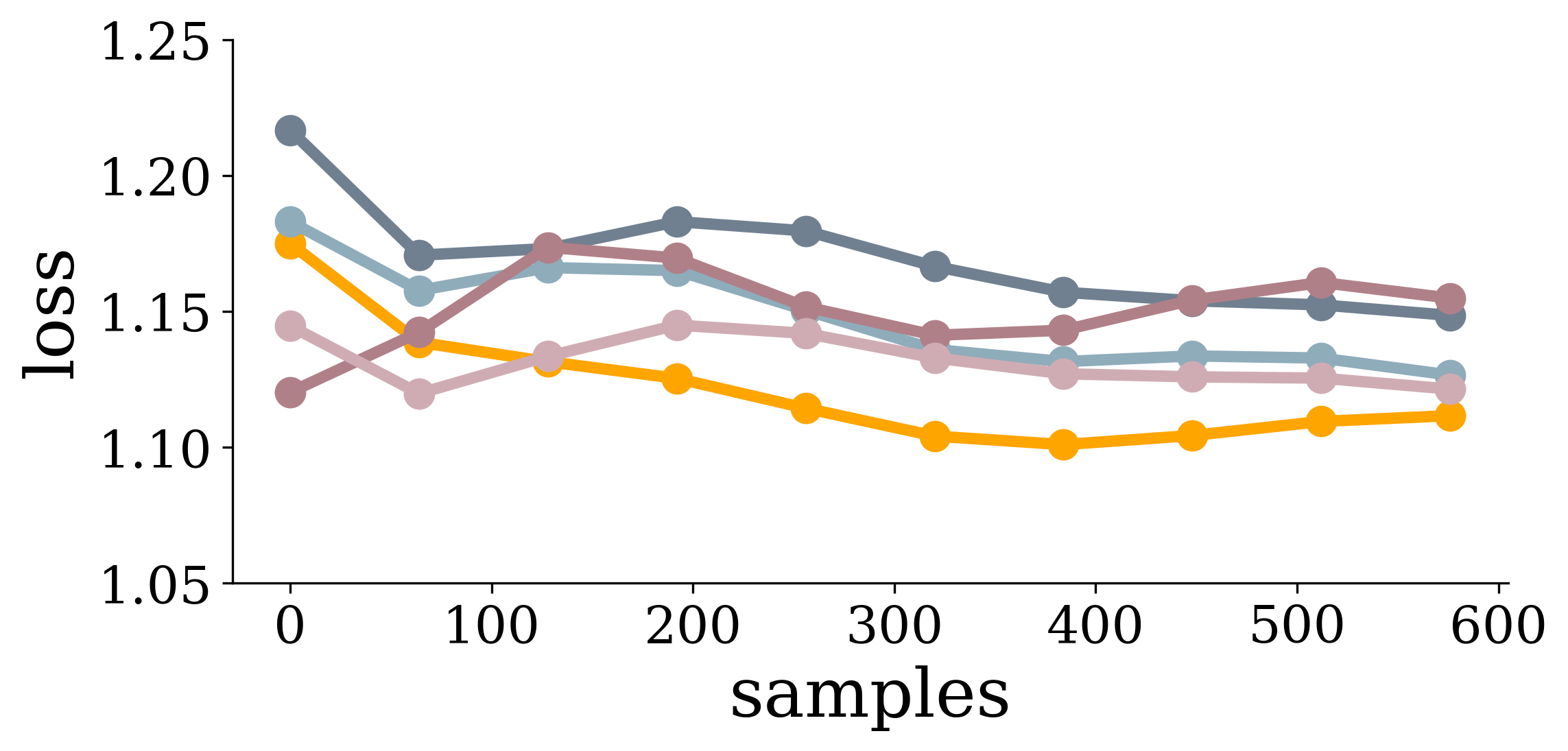}
    \end{minipage}
    \hfill
    \begin{minipage}{\linewidth}
    \caption{Overcooked: we use the human-human dataset from~\cite{Carroll2019OnTU}. For each of the \(5\) layouts in the dataset, we compute the loss of adapting to each new testing partner, and plot the average loss over all the testing partners. For each new partner we have around 1k samples of their actions.}
    \label{fig:overcooked}
    \end{minipage}
\end{figure}

Next, we study the task of Overcooked, a two-player game with the goal of working together to cook and serve food to customers. Players each control a separate avatar in 2D layouts (Figure~\ref{fig:overcooked}), where ingredients and kitchenware are scattered in different locations. Players cannot occupy the same cell, and must interact with the kitchen objects in a certain order to score points (place an onion in the stove, wait, put the cooked onion on a plate, and serve the plate). Of the \(5\) layouts, {\em Cramped Room} is the simplest. {\em Asymmetric Advantages} and {\em Forced Coordination} split the partners into two islands. Lastly, {\em Coordination Ring} and {\em Counter Circuit} have narrow passages that prevent the partners from passing each other.

We test our method on a dataset of human-human demonstrations~\cite{Carroll2019OnTU}, consisting of data for each of the \(5\) layouts shown in Figure~\ref{fig:overcooked}, with around \(8\) training and \(8\) testing pairs of humans per layout (\(39\) training pairs and \(37\) testing pairs in total). The roles of the players are not symmetric; for each demonstration pair we treat player \(0\) as the expert and player \(1\) as the partner. The state is a feature vector of size \(62\), and there are \(6\) available actions for each player.

In Figure~\ref{fig:overcooked}, we show the results of offline evaluation on each of the five layouts from the dataset. Similar to before, we plot the negative log-likelihood loss over the held-out expert data in the test dataset.
The demonstration for each human-human pair in the dataset contains about \(1000\) timesteps on average, which means the total available sample size is limited. Nevertheless, we can see our low-rank partner approach performs well, by inferring the strategy of the partner and predicting the expert actions that can complement this partner. 

\begin{table}[htbp]
    \caption{Reward in Overcooked before/after adapting to test partners, averaged over the \(5\) layouts shown in Figure~\ref{fig:overcooked}.}
    \label{tab:overcooked_static}
    \centering
    \begin{tabular}{c|ccccc}
         \toprule
         &
         {\bfseries lrp} & 
         {\bfseries mt} &
         {\bfseries mod} &
         {\bfseries maml} &
         {\bfseries lt} \\

         \midrule
         before & \(21 \pm 2\) & \({\bm{23}} \pm 2\) & \(17 \pm 2\) & \(7 \pm 1\) & \(21 \pm 2\) \\
         after & \({\bm{24}} \pm 2\) & \(21 \pm 2\) & \(19 \pm 2\) & \(9 \pm 1\) & \(22\pm 2\) \\
         \bottomrule
    \end{tabular}
\end{table}

Next, we look into static evaluation in Table~\ref{tab:overcooked_static}, where we examine the environment rewards attained by the policies when playing with new static partners (averaged over all \(5\) layouts). However, we are given only a dataset of demonstrations, and not the partner policies themselves. As a workaround, we first run Behavioral Cloning over the demonstration dataset to obtain a set of BC partner policies, which we then use to evaluate with our adaptive policies. The rewards attained using this approach are generally poorer because the BC partner policies struggle at novel states, which occur often for conditional MAIL. Still, the results from static evaluation align closely with the same trends we observe from offline evaluation, with the low-rank policy exhibiting good adaptation performance.

\section{User Study: Online Adaptation to Humans}
To see if our method translates well to collaboration with humans, we conduct a study with real humans via a web version of Overcooked. Unlike previous experiments that evaluate against offline datasets or static partners, the user study pairs our adaptive policies against (non-stationary) humans.

\subsection{Participants and Procedures}

We recruited \(40\) workers from the crowd-sourcing platform Prolific and \(10\) local participants, for a total of \(50\) users (48\% Female, median age: 28). The participants were paid at the standard minimum wage rate, and participants from Prolific were prescreened to be from United States and Canada to limit the network latency of playing the online Overcooked game in real time. After providing informed consent, our users were provided text instructions along with a video demonstration of how to play the game. They were then asked to answer a 3-question quiz to check that they understand the rules of the game.
Our study is approved by IRB-49406.

\begin{figure}
    \centering
    \includegraphics[width=\linewidth]{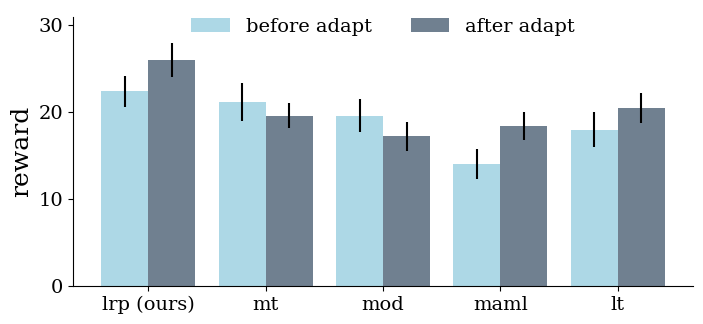}
    \caption{User study: of the \(5\) policy types, {\bf lrp} reaches the highest reward after adapting to human partners.}
    \label{fig:humanstudy}
\end{figure}

We use the {\em Counter Circuit} layout and the adaptive AI policies for Overcooked trained from the previous section. Each participant plays against all \(5\) of the adaptive policies shown in Figure~\ref{fig:humanstudy}, in a randomized order, repeated twice for a total of \(10\) games (\(40\) seconds per game). In each game, the human participant controls Player 2, and the AI agent controls Player 1. The AI agents adapt their policies based only on the human's actions (without access to true or estimated environment rewards).

\noindent
\emph{- Independent Variables:} We vary the policy used to control the AI agent. We choose from one of {\bf lrp}, {\bf mt}, {\bf mod}, {\bf maml}, {\bf lt} in a random order for each user.\\
\emph{- Dependent Measures:} We measure the score (environment reward) attained by the human-AI pair. We report the score from the first interaction ({\em before adapt}) and from the second interaction ({\em after adapt}) for each policy.\\
\emph{- Hypothesis:} Based on the results from offline/static experiments, we hypothesize that {\bf lrp} policy best adapts to the human partners, as measured by {\em after adapt} score.\\
\emph{- Results:} In Figure~\ref{fig:humanstudy} we plot the results of our online user study. Only the {\bf lrp} and {\bf lt} policies attain a reward of \(20\) after adaptation, with the {\bf lrp} indeed giving the highest {\em after adapt} score of \(26 \pm 2\) (statistically significant compared to all the baselines, \(p < 0.05\)). This aligns with the trends we observe from the static/offline evaluation of the Overcooked game, in which the {\bf lrp} also adapted well to new partners.

\section{Conclusion}

\noindent \textbf{Summary.} We study the problem of adapting to new partners in multi-agent tasks, under the imitation learning setting of predicting the expert's actions from the partner's actions.
We formalize the problem setting, learning objective, and the different evaluation metrics for this framework of conditional multi-agent imitation learning.
To address the challenges with learning from limited data, we then propose a low-rank tensor decomposition approach using Tensor Trains. Using a low-rank prior, our model infers the strategy of a new partner and predicts the corresponding expert actions based on the estimated strategy. We describe how to scale up Tensor Trains by parameterizing their functional variant with neural networks, and demonstrate their ability to adapt to new partners on a variety of environments spanning offline, static, and online evaluation. We test on a collaborative bandit task, a continuous-action particle task, the Hanabi game with \(4\)-players, and finally the game of Overcooked in a user study. Our work promotes the novel framework of conditional multi-agent imitation learning, and establishes a promising approach for this framework through incorporating low-rank structure.

\noindent\textbf{Limitations and Future Work.} 
Our model conditions on a new partners actions, but does not actively steer the interaction, e.g., in a Bayesian active learning style. Exploring more active adaptation methods can lead to better convergence guarantees, which is important if a new partner defaults to some non-informative behavior. Moreover, for the non-human experiments, we currently generate the set of partners using self-play with random seeds. Incorporating techniques from literature of training diverse sets of agents can improve the adaptation to new partners.

\section{Acknowledgment}

The authors would like to acknowledge NSF awards 2006388 and 2125511, the Air Force Office of Scientific Research, and the Office of Naval Research.

\bibliographystyle{plain}
\bibliography{ref}

\clearpage
\appendix

We provide more details on the training and testing setup of conditional MAIL. We assume the multi-agent environment has \(n\) agents indexed \(0\) to \(n-1\), and the ego-agent is agent \(0\).

\SetKwComment{Comment}{/* }{ */}
\SetKwInput{KwInput}{Input}
\SetKwInput{KwOutput}{Output}
\SetKwInput{KwReturn}{Return}

\begin{algorithm}
\caption{Training on diverse trajectories}\label{alg:train}
\KwInput{A dataset \(\mathcal{D}\) of \(k\) joint trajectories. Each joint trajectory is collected from a different group of \(n\) agents that have converged with each other.}
\KwOutput{A joint policy \(\pi_\psi\) that will be used as an adaptive policy, with trainable parameters \(\psi\).}

Initialize a joint policy \(\pi_{\theta,\psi}\)\;
\While{training}{
    \For{$j \gets 1,k$}{
        \(\theta \gets \theta + \nabla_\theta \sum_{\mathcal{D}_j} \sum_{i=0}^{n-1} \log \pi_{\theta,\psi} (a^i | s, \psi_j)\)\;
        \(\psi \gets \psi + \nabla_\psi \sum_{\mathcal{D}_j} \sum_{i=0}^{n-1} \log \pi_{\theta,\psi} (a^i | s, \psi_j)\)\;
    }
}

\KwReturn{\(\pi_\psi\)}
\end{algorithm}

In Algorithm~\ref{alg:train}, we have a dataset of \(k\) joint trajectories, where each joint trajectory corresponds to the behavior of a group of \(n\) agents playing in the environment. We assume that the agents have converged to each other, so the \(k\) joint trajectories can be thought of as data from different equilibria of the \(n\)-player environment. Our goal as the ego-agent is to produce the action for player \(0\) (i.e. \(a^0\)). To do so we learn a joint policy \(\pi_{\theta, \psi}\) that has the policy parameters \(\theta\) as well as the partner-strategy parameters \(\psi\) that is different based on the index of the joint trajectory (i.e. the partner ID, or the equilibria ID). This formulation is agnostic to the algorithm used (e.g. {\bf lrp}, {\bf mt}, {\bf mod}), but concretely for {\bf lrp} we note that \(\theta\) corresponds to \(g_1\) and \(\psi\) corresponds to \(g_2\) in Equation~\ref{eq:logits}.

\begin{algorithm}
\caption{Adapting to new partners}\label{alg:test}
\KwInput{A new group of partner policies \(\pi^{1:n-1}\), and the policy \(\pi_\psi\) from Algorithm~\ref{alg:train}.}

Re-initialize partner-strategy parameters \(\psi\)\;
\While{environment steps}{
    \(a^0 \gets \pi_\psi(a^0|s)\)\;

    \For{$i \gets 1,n-1$}{
        \(a^{i} \gets \pi^{i}(a^i|s)\)\;
    }
    \(\psi \gets \psi + \nabla_\psi \sum_{i=1}^{n-1} \log \pi_\psi (a^i | s, \psi_0)\)\;
    \(s \gets T(\cdot | s, a^{0:n-1})\)\;
}
\end{algorithm}

At test time we need to coordinate with one new group of partners. The group has \(n-1\) partner policies for the \(n-1\) roles ranging from \(\pi^1\) to \(\pi^{n-1}\), and the goal of our ego-agent policy is to produce the actions \(a^0\) for agent \(0\). Our ego-agent policy \(\pi_\psi\) is in fact a joint policy: its predictions for the other \(n-1\) actions will be used to finetune its parameters \(\psi\), and only its predictions for action \(a^0\) will be used to step through the environment (combined with the actions \(a^{1:n-1}\) from the other partner policies).

\end{document}